\documentclass[11pt,a4paper]{article}
\usepackage{times,latexsym}
\usepackage{url}
\usepackage[T1]{fontenc}
\usepackage[dvipsnames]{xcolor}
\usepackage{xcolor}

\definecolor{blue_x}{RGB}{121, 173, 220}
\definecolor{orange_x}{RGB}{250, 192, 159}
\definecolor{yellow_x}{RGB}{254, 238, 147}
\definecolor{beige_x}{RGB}{252, 245, 199}
\definecolor{green_x}{RGB}{173, 247, 182}

\usepackage[acceptedWithA]{tacl2021v1}

\usepackage{xspace,mfirstuc,tabulary}

\newif\iftaclinstructions
\taclinstructionsfalse 
\iftaclinstructions
\renewcommand{\confidential}{}
\renewcommand{\anonsubtext}{}
\newcommand{\instr}
\fi

\iftaclpubformat 

\else

\fi

\usepackage[utf8]{inputenc}

\usepackage{microtype}
\usepackage{graphicx}
\usepackage{todonotes}

\usepackage{url}
\usepackage{graphicx}
\usepackage{wrapfig}
\usepackage{amsmath}
\usepackage{amssymb}
\usepackage{color,xcolor,colortbl}
\usepackage{enumitem}
\usepackage{bm,bbm}
\usepackage{booktabs}
\usepackage{mathtools}
\usepackage{array}
\usepackage{subcaption}
\usepackage{pbox}
\usepackage{pifont}
\usepackage{multirow}
\usepackage[scaled=0.7]{beramono}
\usepackage{tabularx}

\usepackage{arydshln}
\definecolor{symptom}{rgb}{1, 0, 0}
\definecolor{disease}{rgb}{1, 0, 0}
\definecolor{complication}{rgb}{1, 0, 0}

\definecolor{test}{rgb}{0.93, 0.53, 0.18}
\definecolor{test-goal}{rgb}{0.93, 0.53, 0.18}
\definecolor{test-result}{rgb}{0.93, 0.53, 0.18}
\definecolor{test-implication}{rgb}{0.93, 0.53, 0.18}

\definecolor{procedure}{rgb}{0.0, 0.5, 0.0}
\definecolor{medicine}{rgb}{0.0, 0.5, 0.0}
\definecolor{treatment-goal}{rgb}{0.0, 0.5, 0.0}
\definecolor{treatment-result}{rgb}{0.0, 0.5, 0.0}

\definecolor{emergency-instruction}{rgb}{0.01, 0.28, 1.0}
\definecolor{upcoming-test}{rgb}{0.01, 0.28, 1.0}
\definecolor{upcoming-treatment}{rgb}{0.01, 0.28, 1.0}
\definecolor{upcoming-specialist}{rgb}{0.01, 0.28, 1.0}
\definecolor{upcoming-pcp}{rgb}{0.01, 0.28, 1.0}
\definecolor{patient-medication}{rgb}{0.01, 0.28, 1.0}
\definecolor{patient-diet}{rgb}{0.01, 0.28, 1.0}
\definecolor{patient-other}{rgb}{0.01, 0.28, 1.0}

\title{PaniniQA: Enhancing Patient Education Through

Interactive Question Answering} 

\author{
Pengshan Cai \thanks{* indicates equal contribution} $^1$, 
Zonghai Yao \footnotemark[1] $^1$, 
\bf{Fei Liu}$^2$, 
\bf{Dakuo Wang}$^3$,
\bf{Meghan Reilly}$^4$,
\bf{Huixue Zhou}$^5$ \\ 
\bf{Lingxi Li}$^1$, 
\bf{Yi Cao}$^1$, 
\bf{Alok Kapoor}$^4$, 
\bf{Adarsha Bajracharya}$^4$,
\bf{Dan Berlowitz}$^6$, 
\bf{Hong Yu}$^{1, 4, 6}$\\
University of Massachusetts, Amherst$^1$, 
Emory University$^2$, 
Northeastern University$^3$\\
UMass Chan Medical School$^4$,
University of Minnesota$^5$, 
University of Massachusetts, Lowell$^6$\\
{\{pengshancai, zonghaiyao, lingxili, yicao, hongyu\}@umass.edu},
fei.liu@emory.edu\\ 
d.wang@neu.edu,
Meghan.Reilly@umassmed.edu, zhou1742@umn.edu \\
{\{Alok.Kapoor, adarsha.Bajracharya\}@umassmemorial.org},
Dan\_Berlowitz@uml.edu
}

\date{}

\begin{document}

\maketitle

\begin{abstract} 
Patient portal allows discharged patients to access their personalized discharge instructions in electronic health records (EHRs). However, many patients have difficulty understanding or memorizing their discharge instructions \cite{zhao2017barriers}. In this paper, we present \textbf{\emph{PaniniQA}}, a \textbf{pa}tient-ce\textbf{n}tr\textbf{i}c i\textbf{n}teract\textbf{i}ve question answering system designed to help patients understand their discharge instructions.  
PaniniQA first identifies important clinical content from patients' discharge instructions and then formulates patient-specific educational questions. 
In addition, PaniniQA is also equipped with answer verification functionality to provide timely feedback to correct patients' misunderstandings. 
Our comprehensive automatic \& human evaluation results demonstrate our PaniniQA is capable of improving patients' mastery of their medical instructions through effective interactions\footnote{Our data and code are released at \url{https://github.com/pengshancai/PaniniQA}}.

\end{abstract}

\section{Introduction}
\label{sec:intro}

Limited patient understanding of their medical conditions can lead to poor self-care at home. Upon hospital discharge, physicians often provide discharge instructions to aid in patients' recovery and disease self-management~\cite{federman2018challenges}. 
However, some patients may have difficulty understanding and memorizing instructions due to low health literacy, limited memory, or an absence of supervision. 
For example, research shows that patients only retain a minimal amount of information from discharge instructions, with an immediate forgetting rate of up to 80\%~\cite{kessels2003patients,richard2017communication}. 
Further, when instructions are misinterpreted by patients, there is often a lack of corrective intervention. 
Limitations in a patient's understanding of their medical conditions hinder their prospects of recovery.
It is imperative to investigate new methods of patient education to enhance health outcomes.

\begin{figure}[t]
\centering
\includegraphics[width=3.0in]{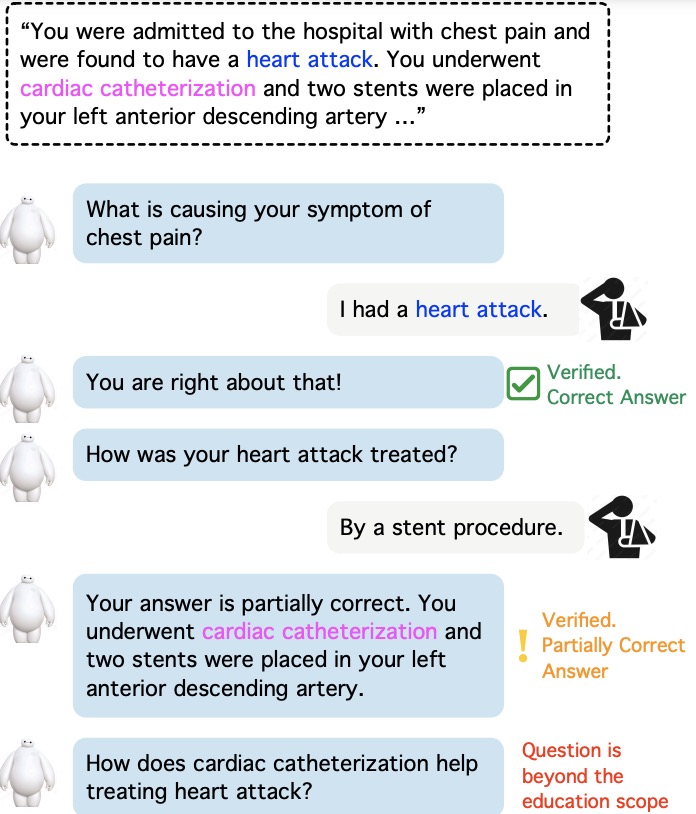}
\caption{An illustration of PaniniQA, our interactive question-answering system for patient education. It generates questions from discharge instructions, helping patients understand their health conditions through interactive question answering. An answer verification module confirms correct responses or expands feedbacks on partially correct ones. The final turn shows a GPT-generated question. Its answer is absent from the discharge instruction and it is deemed inappropriate for patient education.}
\vspace{-0.1in}
\label{fig:demo}
\end{figure}

In this study, we explore a novel method inspired by \emph{Dialogic Reading}~\cite{Whitehurst2002dialogic} to educate patients through interactive question-answering. Dialogic Reading actively involves patients in the learning process by following the PEER sequence: Prompt, Evaluate, Expand, and Repeat, which enables patients to engage in a meaningful dialogue, further strengthening their understanding and retention of the material. As illustrated in Figure~\ref{fig:demo}, our dialog agent asks questions about key aspects of discharge instructions and encourages patients to read and understand the instructions to provide accurate answers thoroughly.

Crafting questions that effectively meet educational objectives is challenging~\cite{boyd-graber-borschinger-2020-question,dugan-etal-2022-feasibility}. A suitable question should be based on the patient's discharge instruction and aim to improve their understanding of health conditions, such as ``\emph{What was the probable cause of your chest pain?}''. Conversely, the question ``\emph{How does cardiac catheterization help treat a heart attack?}'' illustrated in Figure~\ref{fig:demo}, may exceed the education scope, as it is unanswerable or requires knowledge beyond the provided discharge instruction. Such questions are considered unsuitable for patient education.

We introduce new question-generation methods that draw on the advancements of LLMs~\cite{brown2020language,ouyang2022training,openai2023gpt4}. Utilizing OpenAI's GPT-3.5 model, we generate informative questions from discharge instructions. Further, \emph{we combine LLMs with medical event and relation extraction to constrain the model}, producing questions that target salient medical events identified in the discharge instructions. We create a new dataset with expert-annotated medical events and relations for discharge instructions from the MIMIC-III~\cite{johnson2016mimic} database. While earlier efforts have annotated events that physicians would discuss during patient handoff~\cite{pampari-etal-2018-emrqa,lehman-etal-2022-learning}, our focus is on identifying pairs of medical events with correlational or causal relationships. By posing questions about one event, we guide patients toward the other as potential answers.

Our system further incorporates an answer verification module to provide instant patient feedback. When patients give correct answers, the bot confirms them, reinforcing their understanding. If answers are incorrect or partially correct, the bot clarifies misunderstandings and provides additional information. Extensive automatic and human evaluations demonstrate the efficacy of our question-generation methods and show that PaniniQA holds great promise for promoting patient education.
To summarize, our research contributions are as follows.

\begin{itemize}[topsep=3pt,itemsep=-1pt,leftmargin=*]

\item[$\diamond$] We explore a new way of educating patients regarding their health conditions through interactive question-answering. Our approach aligns with the P.E.E.R. dialogic reading theory that promotes patients' active participation in comprehending medical events. 

\item[$\diamond$] We compare questions generated using OpenAI's GPT-3.5 model, our enhanced method with medical event extraction, and human-written questions tailored for patient education. We meticulously evaluated all questions, answers, and patients' educational outcomes.

\item[$\diamond$] Through comprehensive human evaluations, we demonstrate that PaniniQA holds promise for patient education. Future work includes controlling the difficulty of questions, prioritizing questions given patients' health literacy, and enabling interactive learning of medical concepts.

\end{itemize}

\section{Related Work}
\label{sec:related}

There is a growing need to improve patients' understanding regarding their hospital experiences~\cite{federman2018challenges,Weerahandi2018ym, kwon2022medjex}. Lack of understanding can result in non-adherence to discharge instructions and readmission to the hospital due to poor self-care at home. Previous research has attempted to generate hospital course summaries for patients using lay language~\cite{di-eugenio-etal-2014-patientnarr,acharya-etal-2018-towards,adams-etal-2021-whats,cai-etal-2022-generation,Hartman2022ax, adams-etal-2022-learning}. This paper goes a step further by utilizing interactive question answering to communicate essential medical events from discharge instructions to patients, thus enhancing their understanding and retention of the material.

Our proposed method differs from existing clinical question-answering studies in several aspects. Most clinical QAs are designed to satisfy individuals' information needs, with questions modeled after those that can be asked by physicians~\cite{pampari-etal-2018-emrqa,jin-etal-2019-pubmedqa,raghavan-etal-2021-emrkbqa,lehman-etal-2022-learning}. These systems focus on improving the accuracy of their answers~\cite{soni-roberts-2020-evaluation,rawat-etal-2020-entity,yue-etal-2020-clinical,cliniqg4qa2020}. In contrast, our goal is to educate patients and prompt them with questions that will enhance patients' understanding of their doctors' visits. A successful QA system should be \emph{comprehensive} and \emph{exhaustive}, asking all relevant questions and prioritizing them based on the patient's medical history and health literacy.

Successful patient education requires effective questioning~\cite{pylman2020}. Particularly, question generation has been studied using template-based~\cite{heilman-smith-2010-good,chali2015,fabbri-etal-2020-template} and neural seq2seq models~\cite{du-cardie-2017-identifying,duan-etal-2017-question,kim2018improving,sultan-etal-2020-importance,shwartz-etal-2020-unsupervised}. Instruction-tuned LLMs have demonstrated exceptional abilities in conversing with humans~\cite{brown2020language,Sanh2021,ouyang2022training,chowdhery2022palm,longpre2023flan}. However, most research has been conducted using CommonCrawl, Wikipedia, and other generic texts. Considering the factuality issues of neural language models~\cite{maynez-etal-2020-faithfulness,pagnoni-etal-2021-understanding}, question generation in the medical domain remains challenging.

Learning through conversation can improve education outcomes~\cite{golinkoff2019language,zhang-etal-2020-dialogpt,cai-etal-2022-learning,yao-etal-2022-ais, yao-etal-2022-ais, xu-etal-2022-fantastic}. \emph{Dialogic Reading}~\cite{Whitehurst2002dialogic,mol2008added,lever2011discussing} has demonstrated that engaging children in a guided conversation with parents while reading storybooks can significantly enhance their learning outcomes. While engaging physicians in high-quality conversations may not always be feasible, the use of question answering facilitated by a chatbot could be a valuable means of helping patients acquire a deeper understanding of their health conditions.

\section{Question Answering in the GPT Era} 
\label{sec:gpt}

Large language models (LLMs) such as ChatGPT have led to significant advancements in generative AI~\cite{brown2020language,Sanh2021,chowdhery2022palm,longpre2023flan,openai2023gpt4,wang-etal-2023-umass}. Fine-tuning neural models on specific tasks often yields superior results. Furthermore, LLMs acquire emergent abilities through instruction tuning and reinforcement learning using human feedback~\cite{ouyang2022training}. This allows them to generalize to new tasks effectively. Common human-LLM interactions include (a) \emph{zero-shot prompting}, where users provide a prompt for the LLM to complete, and (b) \emph{in-context learning}, where users give task examples and ask the LLM to solve a new case, potentially involving a multi-step reasoning process~\cite{wei2022chain}. In this study, we focus on zero-shot prompting to assess the LLM's ability to comprehend discharge instructions.

LLMs possess vast world knowledge, and their performance on knowledge-intensive tasks correlates with training data and model size~\cite{bommasani2022opportunities}. However, it remains unclear whether LLMs have enough domain knowledge to facilitate patient education. For example, GPT-3, with its 175 billion parameters, is trained on general data sources such as Common Crawl, WebText2, Books, and Wikipedia~\cite{brown2020language}. Yet, the model still generates factually inconsistent errors within their output. Our study presents an initial evaluation of GPT models' potential in interactive patient education. Following the P.E.E.R. framework of dialogic reading, we employ GPT models to perform the following tasks:

\vspace{0.04in}
\noindent \textbf{\emph{Question Generation.}}\label{sec:question-gen}\quad 
We use OpenAI's GPT-3.5 model (\texttt{text-davinci-003}) to generate informative questions from a discharge instruction. The questions aim at helping patients understand crucial medical events. Our prompt is ``{\emph{Generate N questions to help the patient understand crucial medical events in the above discharge instruction.}}'' Similar to a teacher designing exam questions, we anticipate the GPT model to produce a set of questions all at once rather than incrementally. The questions must collectively cover the salient events identified in the discharge instruction while minimizing redundancy.

\vspace{0.04in}
\noindent \textbf{\emph{Answer Verification.}}\label{sec:answer-veri}\quad
Useful feedback is essential for improving patient comprehension of the material. To perform this task, we prompt the GPT model with ``\emph{As a physician, your goal in the conversation is to help your patient better understand the discharge instructions before they leave the hospital.}'' Utilizing OpenAI's API, we also provide the original discharge instruction, interaction history, and current question-answer pair as key-value pairs for the model. We then instruct the model to ``\emph{verify if the patient's answer is correct, incorrect, or partially correct, and generate a suitable response to improve the patient's comprehension of this question.}'' We empirically compared two GPT models, \texttt{text-davinci-003} and \texttt{gpt-3.5-turbo} (ChatGPT), and selected ChatGPT for answer verification as it is optimized for chat and generally produces higher quality responses.

\begin{table*}[t]
\setlength{\tabcolsep}{11.5pt}
\renewcommand{\arraystretch}{1.1}
\centering
\begin{scriptsize}
\begin{tabular}{ll}
\toprule
\textsc{{Binary Relation of Events}} & \textsc{{Question Templates}} \\ 
\midrule
\textcolor{symptom}{Symptom} /  \textcolor{disease}{Disease} & \textbf{Q}: What is the cause of your symptom [Symptom]? \textbf{A}: [Disease] \\
\textcolor{test}{Test} / \textcolor{test-goal}{Test goal} & \textbf{Q}: What is the goal of test [Test]? \textbf{A}: [Test goal] \\
\textcolor{test}{Test} / \textcolor{test-result}{Test result} & \textbf{Q}: What is the result of test [Test]? \textbf{A}: [Test result] \\
\textcolor{test}{Test} /  \textcolor{test-implication}{Test implication} & \textbf{Q}: What does test [Test] imply? \textbf{A}: [Test implication] \\
\textcolor{procedure}{Procedure} (or  \textcolor{medicine}{Medicine}) /  \textcolor{treatment-goal}{Treatment goal} & \textbf{Q}: What is the goal of treatment [Procedure] (or [Medicine])? \textbf{A}: [Treatment goal] \\
\textcolor{procedure}{Procedure} (or \textcolor{medicine}{Medicine})  / \textcolor{treatment-result}{Treatment result} & \textbf{Q}: What is the result of treatment [Procedure] (or [Medicine])? \textbf{A}: [Treatment result] \\
\bottomrule
\end{tabular}
\end{scriptsize}
\vspace{-0.05in}
\caption{Expert-written question templates are used to generate a question from each binary relation. This method enables us to create targeted questions about salient medical events. By posing questions about one event, we guide patients towards the other as potential answers. The placeholders are to be replaced with medical events detected from discharge instructions.} 
\label{tab:evt_type_templates}
\vspace{-2mm}
\end{table*}

\section{Extracting Salient Medical Events}
\label{sec:in-house}

In this section, we present our question-answering system that emphasizes identifying salient medical events and their relations. We generate targeted questions using them and apply the same answer verification module described previously.

A typical discharge instruction includes \textbf{\emph{Visit Recap}}, which recaps a patient's clinical visit, including symptoms, diagnoses, treatments, and test results. Patients are expected to understand the relationships among these medical events, such as how the treatment \emph{ERCP} relates to \emph{cholangitis} as illustrated in Table~\ref{tab:qdemo} (\emph{top}). \textbf{\emph{Detailed Instructions}} include medication and aftercare instructions (\emph{bottom}). They may be easy to understand but contain trivial details that patients may overlook, potentially hindering their self-care at home.
We propose automatically extracting key medical events and relations from them (\S\ref{sec:event_identification}). Given their unique characteristics, we apply two distinct information extraction and question generation strategies for \emph{Visit Recap} and \emph{Detailed Instructions} to produce targeted questions (\S\ref{sec:question_generation}).

\begin{table}[]
    \centering
    \begin{scriptsize}
    \begin{tabularx}{0.48 \textwidth}{X}
    \toprule
    \textsc{QG - Visit Recap} \\
    \midrule
    \rowcolor{gray!10}
    \emph{You were found to have an infection of your bile ducts called \textcolor{disease}{cholangitis} . You had a procedure called an \textcolor{procedure}{ERCP} where a stent was placed to relieve the obstruction ...} \\
    \textbf{Relation}: \textcolor{procedure}{ERCP} (Procedure) -- \textcolor{treatment-goal}{cholangitis} (Treatment Goal) \\
    \textbf{Question}: What is the goal of treatment \textcolor{procedure}{ERCP}? \\
    \midrule
    \textsc{QG - Detailed Instructions} \\
    \midrule
    \rowcolor{gray!10}
    \emph{We made the following changes to your medication regimen: 1. We started you on a new medication called Toprol XL 25mg by mouth \textcolor{patient-other}{twice a day} ... }\\
    \textbf{Event}: \textcolor{patient-other}{twice a day} (Medicine Frequency) \\
    \textbf{Question}: How often should Toprol XL be taken?\\
    \bottomrule
    \end{tabularx} 
    \end{scriptsize}
    \caption{Question generation (QG) from a medical event (bottom) or a binary relation of events (top).}
    \label{tab:qdemo}
    \vspace{-5mm}
\end{table}

\subsection{Event and Relation Identification} 
\label{sec:event_identification}

Key event and relation identification are conducted on \textbf{\emph{Visit Recap}}. Event identification is framed as a \emph{sequence labeling} task, where we assign a label to each token of the discharge note, representing its event type. We define 11 event types in this study, detailed in Table~\ref{tab:anno-schema}, including symptoms, diseases, complications, tests, test goals/results/implications, procedures, medicines, treatment goals and results. We fine-tune pre-trained sequence labeling models on our dataset, optimizing the cross-entropy loss of gold standard labels.

Relation identification is framed as a \emph{sequence classification} task. We focus on binary relations consisting of two medical events. We evaluate all pairwise combinations of identified medical events as candidates, provided their event types align with the six event relations defined in Table~\ref{tab:evt_type_templates}. Special tokens are inserted before and after each identified event to indicate both its position and event type.\footnote{E.g., the sentence ``\emph{You were admitted for diverticulitis and treated with antibiotics}'' was modified as ``\emph{You were admitted for <dsyn> diverticulitis </dsyn> and treated with <medi> antibiotics </medi>}, where the special tokens \emph{<dsyn>} and \emph{</dsyn>}'' indicates the start and end position of this event, and \emph{dsyn} reflects the event belongs to the category \emph{Disease}.} The sequence, enhanced with special tokens, is fed into a sequence classification model to predict a binary label, where 1 indicates a relation between the two events, and 0 otherwise. We fine-tune pre-trained sequence classification models on our dataset (\S\ref{sec:annotation}) by optimizing the cross-entropy loss for gold-standard labels.

We perform key event identification on \emph{\textbf{Detailed Instructions}} using a different tool, as they contain medication and aftercare specifics that patients might overlook. We use an existing high-performing medical NER system to extract medical entities.\footnote{\url{https://pypi.org/project/Bio-Epidemiology-NER/}} This model was pre-trained on the MACCROBAT dataset~\cite{caufield2019comprehensive} and can identify 84 biomedical entities within clinical narratives. We limit the model to identify 7 entity types: \emph{Medicine Dosage}, \emph{Medicine Frequency}, \emph{Medicine Duration}, \emph{Medication Name}, \emph{Sign \& Symptom}, \emph{Diagnostic Procedure}, \emph{Upcoming Appointment}. Relation identification is not performed on detailed instructions.

\subsection{Question Generation}
\label{sec:question_generation}

\vspace{0.04in}
\noindent\textbf{Visit Recap.}\quad
We generate a question from each identified binary relation. Different relation types are mapped to specific questions using templates provided by physicians according to their domain knowledge (see Table~\ref{tab:evt_type_templates}). Using a template-based approach allows us to create questions targeting salient medical events. By asking questions about one event, we guide patients towards the other as potential answers.

\vspace{0.04in}
\noindent\textbf{Detailed Instructions.}\quad
We generate a question for each identified medical entity by creating a fill-in-the-blank question, which is then converted into a natural language question using the GPT model. An example is shown in Table~\ref{tab:qdemo}. Although cloze-style questions can serve educational purposes, we want to prevent patients from using string matching to find answers. Instead, natural language questions require patients to have a deeper understanding of the discharge note, thus fulfilling our education objective. When selecting medical entities as triggers, we prioritize four categories: \emph{Medicine Dosage}, \emph{Medicine Frequency}, \emph{Medicine Duration}, and \emph{Upcoming Appointment}, as they are informative and better guide patient comprehension. To convert a cloze-style question into a natural question, we provide this prompt to the GPT model: {\emph{[Fill-in-the-Blank Sentence] Generate a simple question targeting the blank in the above sentence.}}

\section{Data Annotation}
\label{sec:annotation}

We seek to annotate discharge instructions from the MIMIC-III database (v1.4)~\cite{Johnson2016} with key medical events that are important for patients to understand. MIMIC-III is a publicly available repository of de-identified health records of over 40,000 patients collected from the Beth Israel Deaconess Medical Center in Massachusetts. Our aim is to identify text snippets in discharge instructions that correspond to significant medical events, including \emph{symptoms}, \emph{diseases}, \emph{test results}, and \emph{treatments}. 
We annotate not only individual events but also their relationships. They are organized into a \emph{hierarchy} as outlined in the schema shown in Table~\ref{tab:anno-schema}.
Consistent with Lehman et al.~\shortcite{lehman-etal-2022-learning}'s approach, we utilize events and their relationships as \emph{triggers} that prompt the generation of questions.

We recruited five medical experts to create a sizable dataset. They are M.D. students at UMass Chan Medical School and have a high level of language proficiency. 
Each expert is given 150 discharge notes to annotate. 
It is possible to skip some notes due to low text quality. Annotators were also given detailed instructions and examples. We developed a web-based interface to facilitate the annotation process, which has been iteratively improved to meet the needs of this study. Due to budget constraints, we assign one annotator to each discharge note. In total, we completed 458 discharge notes with medical event annotations.

\begin{table}
\setlength{\tabcolsep}{0pt}
\renewcommand{\arraystretch}{1}
\centering
\begin{footnotesize}
\begin{tabular}{l r}
\toprule
\textsc{Medical Events \& Relations} & \textsc{\# of Instances} \\
\midrule
E-1.1 \textcolor{symptom}{Symptom} & 772\\
E-1.2 \textcolor{disease}{Disease} & 541 \\
E-1.3 \textcolor{complication}{Complication} & 135 \\
R-1.1 [\textcolor{symptom}{Symptom}] caused by [\textcolor{disease}{Disease}] & 323 \\
\midrule
E-2.1 \textcolor{test}{Test}  & 216 \\
E-2.2 \textcolor{test-goal}{Test goal} & 17 \\
E-2.3 \textcolor{test-result}{Test result} &  208 \\
E-2.4 \textcolor{test-implication}{Test implication} & 24 \\
R-2.1 [\textcolor{test}{Test}] goal: [\textcolor{test-goal}{Test-Goal}] & 12 \\
R-2.2 [\textcolor{test}{Test}] result: [\textcolor{test-result}{Test-Result}] & 163 \\
R-2.3 [\textcolor{test}{Test}] implication: [\textcolor{test-implication}{Test-Implication}] & 20 \\
\midrule
E-3.1 \textcolor{procedure}{Treatment}\\
\quad E-3.1.1 \textcolor{procedure}{Procedure} & 359 \\
\quad E-3.1.2 \textcolor{medicine}{Medicine}  & 536 \\
E-3.3 \textcolor{treatment-goal}{Treatment goal} & 86 \\ 
E-3.4 \textcolor{treatment-result}{Treatment result} & 239 \\
R-3.1 [\textcolor{procedure}{Treatment}] goal: [\textcolor{treatment-result}{Treatment-Goal}] & 287\\
R-3.2 [\textcolor{procedure}{Treatment} result: [\textcolor{treatment-result}{Treatment-Result}] & 237 \\

\bottomrule
\end{tabular}
\end{footnotesize}
\caption{
A hierarchy of salient medical events. We consider both medical events (E) and their binary relationships (R). 
}
\label{tab:anno-schema}
\vspace{-5mm}
\end{table}

\begin{figure}[t]
\centering
\includegraphics[width=2.8in]{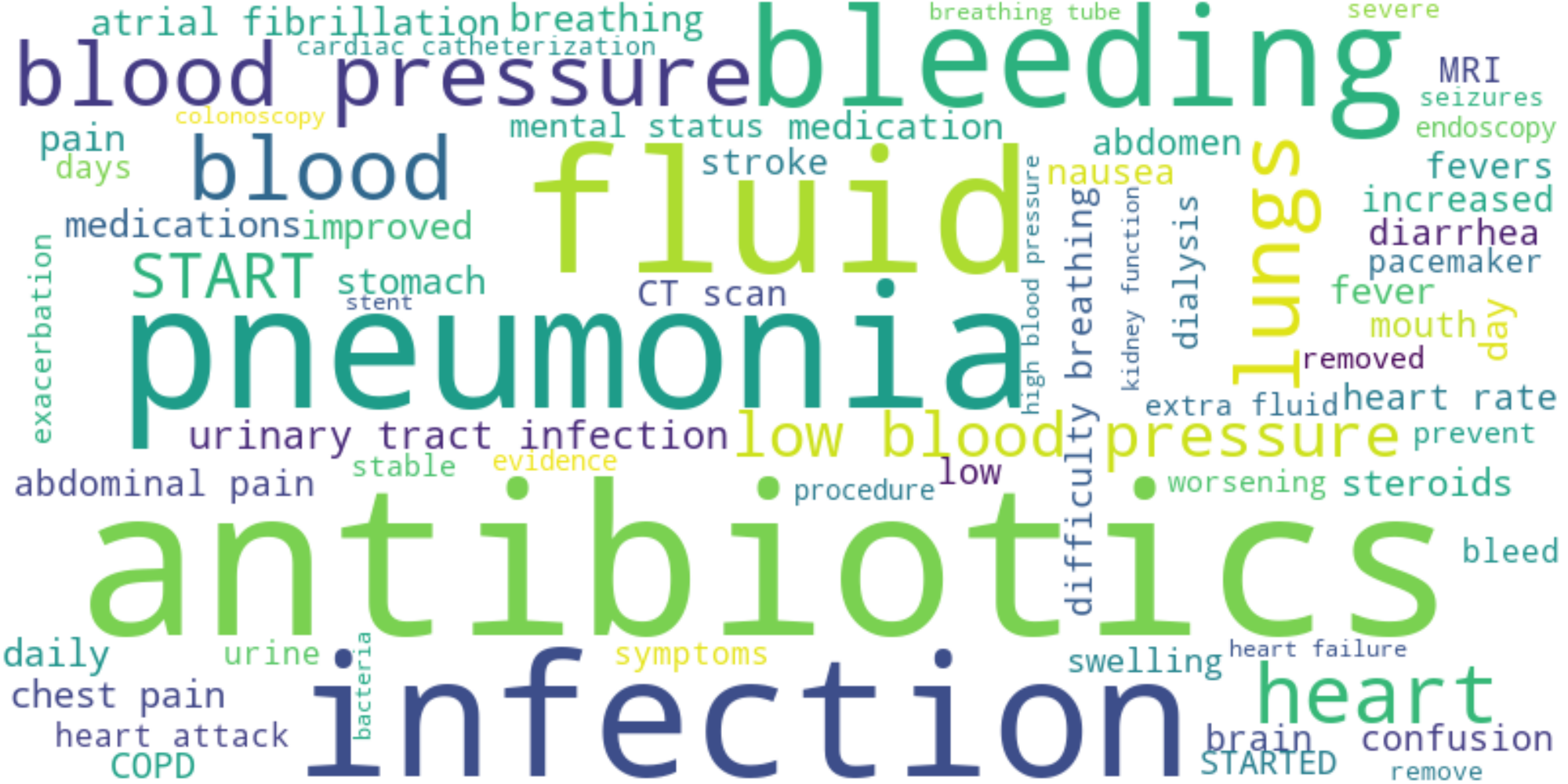}
\caption{Word cloud demonstrating the most frequent medical terminologies and their frequency in our annotations. The sizes of the terminologies refer to their frequency in our dataset. These terminologies are identified from annotated medical events using SciSpacy.}
\vspace{-0.1in}
\label{fig:wordcloud}
\end{figure}

Our annotation consists of two phases. In the first phase, an expert selects text snippets from the discharge instruction corresponding to medical events that the patient needs to understand. Each snippet is assigned a coarse event category, such as a \emph{medical issue}, \emph{laboratory test}, \emph{treatment}. The expert further refines it by assigning a fine-grained event type, resulting in a schema with 11 event types (Table~\ref{tab:anno-schema}). In the second phase, the expert identifies relationships between medical events using a set of 6 pre-defined relationships, such as ``\emph{[\textcolor{black}{Symptom}] ... caused by [\textcolor{black}{Disease}]}.'' We show a distribution of medical events in Figure~\ref{fig:wordcloud}.

A key distinction between our work and earlier dataset curation efforts~\cite{pampari-etal-2018-emrqa,cliniqg4qa2020,lehman-etal-2022-learning} is that the earlier efforts aim to annotate \emph{questions} that physicians would ask during patient hand-off, which may be informal and unanswerable based on the discharge instruction. In contrast, our focus is on annotating \emph{salient medical events} that are essential to patient's understanding of their medical conditions.

We split our annotated data into train / validation / test splits, which contain 338 / 60 / 60 discharge instructions, respectively. 
For relation identification, we use the event pairs from the human-annotated relations as positive relations and all other medical event pairs of compliant types (e.g., the event pair types in Table~\ref{tab:anno-schema}) as negative relations. We collect all negative event pairs\footnote{I.e. No relationship exist between the event pair, in addition, the  types of the two events are restricted by Table~\ref{tab:evt_type_templates}} as negative cases.
Overall, our medical relation dataset contains 2530 / 399 / 332 instances in the train / validation / test set, respectively; 28.7\% instances are positive relations.

\section{Evaluating Information Extraction}
\label{sec:exp}

\begin{table}
\setlength{\tabcolsep}{9pt}
\renewcommand{\arraystretch}{1}
    \centering
    \begin{scriptsize}
    \begin{tabular}{clrrr}
    \toprule
    & \textbf{Pretrained Model} & \textbf{P}(\%) & \textbf{R}(\%) & \textbf{F1}(\%) \\
    \midrule
    \parbox[t]{2mm}{\multirow{4}{*}{\rotatebox[origin=c]{90}{Medical}}} 
    \parbox[t]{2mm}{\multirow{4}{*}{\rotatebox[origin=c]{90}{Events}}} 
    & Bert    &   31.38   & 44.58  &  36.83  \\
    & BioBert    &  40.43 & 51.63 &  45.35   \\
    & PubmedBERT   & 42.70 & 50.12 &  46.11   \\
    & ClinicalRoBERTa   &  \textbf{44.28} & \textbf{54.03} &  \textbf{48.67}   \\
    \midrule
    \parbox[t]{2mm}{\multirow{4}{*}{\rotatebox[origin=c]{90}{Event}}} 
    \parbox[t]{2mm}{\multirow{4}{*}{\rotatebox[origin=c]{90}{Relations}}} 
    & Bert                  & 57.48     & 75.31     & 65.21  \\
    & BioBert               & 73.41     & 80.37     & 76.73 \\
    & PubmedBERT            & 72.56     & 75.31     & 73.91 \\
    & ClinicalRoBERTa          & \textbf{74.28}   & \textbf{82.27}     & \textbf{78.07} \\
    \bottomrule
    \end{tabular}
    \end{scriptsize}
    \vspace{-0.02in}
    \caption{Results of fine-tuning four pretrained models on IE task: medical event extraction (Top) and event-relation identification (Bottom).}
    \label{table:model_compare}
\end{table}

\begin{table}
\setlength{\tabcolsep}{13pt}
\renewcommand{\arraystretch}{0.9}
    \centering
    \begin{scriptsize}
    \begin{tabular}{lccc}
    \toprule
    \textbf{Medical Event} & P (\%) & R (\%) & F1 (\%) \\
    \midrule
    \textcolor{symptom}{Symptom} & 50.8 & 78.2 & 61.6 \\
    \textcolor{disease}{Disease} & 54.3 &  73.5 & 62.5 \\
    \textcolor{disease}{Complication} & 25.0 & 23.5 & 24.2 \\
    \midrule
    \textcolor{test}{Test} & 65.9 & 82.8 & 73.4 \\
    \textcolor{test-goal}{Test goal} & 25.0 & 25.0 & 25.0 \\
    \textcolor{test-result}{Test result} & 36.3 & 30.0 & 32.8  \\
    \textcolor{test-implication}{Test implication} & 16.6 & 16.6 & 16.6 \\
    \midrule
    \textcolor{procedure}{Procedure} & 38.5 & 53.6 & 44.8 \\
    \textcolor{procedure}{Medicine} & 42.3 & 42.3 & 42.3 \\
    \textcolor{treatment-goal}{Treatment goal} & 16.6 & 28.5 & 28.0 \\
    \textcolor{treatment-result}{Treatment result} & 19.0 & 22.8 & 20.7 \\
    \midrule
    Overall & 44.2 & 54.0 & 48.6 \\
    \bottomrule
    \end{tabular}
    \end{scriptsize}
    \caption{Automatic evaluation results of medical event identification per category with ClinicalRoBERTa model.}
    \label{table:evt_idf}
    \vspace{-5mm}
\end{table}

\begin{table}[t]
\setlength{\tabcolsep}{5pt}
\renewcommand{\arraystretch}{1}
    \centering
    \begin{scriptsize}
        \begin{tabular}{lrrr}
        \toprule
        \textbf{Medical Event Relation} & \textbf{P}(\%) & \textbf{R}(\%) & \textbf{F1}(\%) \\
        \midrule\relax
        [\textcolor{symptom}{Symptom}] caused by [\textcolor{disease}{Disease}]             &   81.25   & 79.59     & 80.41      \\
        
        [\textcolor{test}{Test}] goal: [\textcolor{test-goal}{Test-Goal}]                 &   100.0   & 60.0      & 75.0     \\
        
        [\textcolor{test}{Test}] result: [\textcolor{test-result}{Test-Result}]              &   61.54   & 92.31     & 73.85     \\
        
        [\textcolor{test}{Test}] implication: [\textcolor{test-implication}{Test-Implication}]          &   57.14   & 66.67     & 61.54     \\
        
       [\textcolor{procedure}{Treatment}] goal: [\textcolor{treatment-result}{Treatment-Goal}]            &   81.82   & 81.82     & 81.82     \\
       
        [\textcolor{procedure}{Treatment} result: [\textcolor{treatment-result}{Treatment-Result}]         &   70.59   & 85.71     & 77.42      \\
        \midrule
        Overall                         &   74.28   & 82.27     & 78.07     \\
        \bottomrule
        \end{tabular}
    \end{scriptsize}
    \caption{Event-Relation detection per category with ClinicalRoBERTa model.}
    \label{table:rel_idf}
    \vspace{-3mm}
\end{table}

To improve LLMs' ability to generate educationally effective questions for patient education, we designed an \textbf{Information Extraction (IE)} module (medical event/relation identification) to guide question generation. 
We report automatic evaluation results for different IE methods in this section.

\subsection{IE Evaluation Settings}\quad
\noindent We fine-tune four pre-trained language models on our annotated dataset for key medical event and relation identification in Section~\ref{sec:annotation}. These models are obtained from HuggingFace:
1) BERT-large \cite{devlin-etal-2019-bert}; 
2) BioBERT \cite{lee2020biobert}; 
3) PubmedBERT \cite{pubmedbert}; 
4) ClinicalRoBERTa \cite{lewis2020pretrained}; 
All four pre-trained models have the same scale of parameters (345 million). The later three language models were pre-trained on different bio-medical or clinical corpora, thus are better transferable to our patient education task due to the model's level of medical knowledge~\cite{sung2021can, yao2022extracting, yao2022context}.
The models are trained on a single RTX 6000 GPU with 24G memory. The average training time for the relation identification model is around 20 minutes.\footnote{Due to data sparsity, when training both the medical event and relation identification models, we first explore the optimal hyper-parameter set using the validation set. We then combine the validation set into the train set to train our models.}
For evaluation metrics, we report the model's Micro-average precision, recall, and F-1 score.

\vspace{0.04in}
\subsection{IE Evaluation Results}\quad
The performance of four evaluated models is in Table~\ref{table:model_compare}. The results suggest: 
The models pre-trained with biomedical or clinical corpus show better performance than the naive Bert model. 
For both tasks, ClinicalRoberta achieves the best performance, so we report only this model's performance in following category-wise performance analysis.

We further report more fine-grained results of the medical event extraction per category in Table~\ref{table:evt_idf}, and the \emph{Symptom}, \emph{Disease}, \emph{Test}, \emph{Procedure} and \emph{Medicine} categories generally achieve better performance, as we suspect it is due to a more abundant training data. 
Table~\ref{table:rel_idf} shows the fine-grained performance of event-relation identification per category.
The F-1 scores of most relations are around 80\%, implying fair performance. 
The relation \emph{Test goal} achieves 100\% in precision because our test set contains eight \emph{Test goal} instances. 

To explore the generalization ability of the model, we compare the model's performance on seen and unseen medical events during training. Specifically, seen events are events that appear in the training set, while unseen events are not. We observed that 15.21\% of the test instances are unseen medical events. For the medical-event extraction task, the F1 score of seen events is 49.36\%, and the F1 score of unseen events is 44.82\%. For the event-relation identification task, if both event in the event pair are seen events, the model achieves 78.72\% in F1 score, otherwise the performance drop to 74.50\%. This indicates the model only shows slight drop in performance when encountering unseen medical events during training.

\begin{figure*}[t]
\centering
\includegraphics[width=0.9\linewidth]{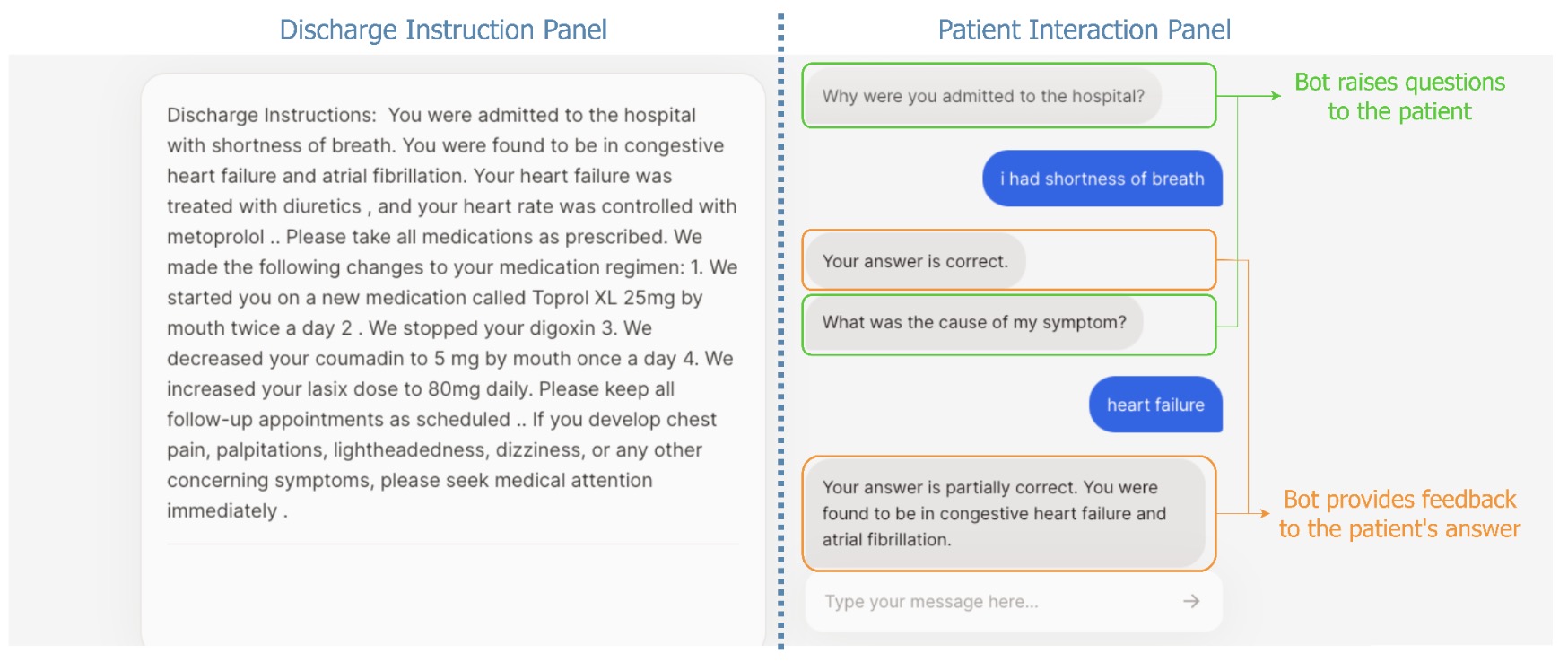}
\vspace{-0.1in}
\caption{
System UI of our human evaluation study: the left panel shows a discharge note (Condition \textit{\textbf{None}}), and the right panel provides the question-answering interactions to the user via a chatbot. The bot can either present only questions (Condition \textit{\textbf{Q}} in \textcolor{green_x}{green boxes}) or plus answer feedbacks (Condition \textit{\textbf{QA}} in \textcolor{orange_x}{orange boxes}).
}
\label{fig:screenshot}
\vspace{-3mm}
\end{figure*}

\section{Evaluating Patient Education}
\label{sec:human_eval}
In order to comprehensively evaluate patient education outcome, We conducted human evaluations from both the patient's and the physician's perspectives, as well as a GPT-4 powered automatic evaluation.  These evaluations focus on two main aspects: 1) The generated question's quality of different models (GPT, GPT+IE, and human ground-truth);  2) The preference of different designs of the interaction experience (None of support, Raising Questions only, and Raising Questions and Verifying Answers).

\subsection{Human Evaluation Settings}

The goal of \textbf{physician evaluation} is to have human domain experts evaluate whether these machine-generated questions are comparable to the human-crafted questions or not. 
To do so, we recruited 3 medical practitioners\footnote{Two licensed physicians and one medical student with hospital internship experience} and their tasks are to read the discharge instructions, and provide qualitative feedback on if these machine-generated questions are educationally effective to the patients; if not, how should they be improved.

The goal of \textbf{patient evaluation} is to have the general public users interact with  and provide ratings on the different combinations of the question-generation models and the interaction designs. We also designed a post-experiment evaluation task (i.e., Cloze Test) to quantitatively measure their understanding outcome.
We recruit 30 human evaluators to participate in our patient education experiment.
All the evaluators have bachelor's degrees but do not have any medical education background.

In our study, we have the following three options for the user interaction experience design:
\begin{enumerate}[itemsep=-0.3mm]
\vspace{-0.3em}
  \item Condition \emph{None}: The evaluator only sees the discharge instruction, no question-answer interaction. This is today's baseline.
  \item Condition \emph{Q}: The evaluator reads the discharge instruction, and interact with the chatbot, which can only ask questions but do not to provide feedback to users' answers.
  \item Condition \emph{QA}: The evaluator reads the discharge instruction, and interact with the chatbot, which can ask questions and provide answer feedback to the user.
\end{enumerate}

The questions asked by the chatbots can come from following three sources:

\begin{enumerate}[itemsep=-0.3mm]
\vspace{-0.3em}
  \item \textit{Human}: Expert-written questions based on discharge instructions. We ask an MD student to read each discharge instruction and write down all questions she would ask a patient about this discharge instruction for patient-education purposes. 
  \item \textit{GPT}: We utilize GPT-3 model to generate a series of questions (at least four) directly from the discharge instruction. Specifically, we use the following prompt: \emph{[Discharge Instruction] Generate at least four questions to help the patient understand crucial medical events in the above discharge instruction.} \footnote{We have tried a collection of prompts for the similar purpose, and do not observe significant differences in the quality of generated questions. We used the chosen prompt as it is naive to understand and leads to more succinct questions. Specifically, we instruct GPT-3 to generate at least four questions to benchmark against the least number of questions from the human annotator. }  
  \item \textit{GPT+IE}: Our question generation model enhances by the information extraction technique described in Sec~\ref{sec:in-house}.
\end{enumerate}

The average number of questions from approach \textit{Human} / \textit{GPT} / \textit{GPT+IE} are 7.5 /  6.17 / 6.1. When combining the variety of interaction designs and question-generation methods, there are five different conditions: 1) \emph{None}; 2) \emph{Q} (\textit{Human}); 3) \emph{QA} (\textit{GPT}); 4) \emph{QA} (\textit{GPT+IE}); 5) \emph{QA} (\textit{Human}).
We perform a within-subject experiment setup, where each of the 30 human evaluators should experience all five conditions using different discharge instructions. In total, we have 150 data points (30 per each condition). The order of the five conditions are shuffled so that each condition appears six times at each of the five orders.

\subsection{Patient Evaluation Measurements}
\label{sec:human_eval_measure}

We use two measurements to evaluate patient's educational outcome and preference.

\noindent1) \textsc{\textbf{Cloze Test}}:  We recruited an MD student to identify 5-7 important medical events that she thinks the patient should be aware of, and replace them with blanks. We use these cloze tests as a post-study evaluation to ask each participant to try their best to fill in the blanks using their memory. 
The more blanks they fill in correctly, the better the patient's education outcome is. We report the participant's accuracy rate as the primary evaluation outcome.

\noindent2) \textsc{\textbf{Preference Ranking}}: We ask evaluators to rank their experience using the following four questionnaire items (Evaluators are allowed to rank two conditions as tied): 
\begin{itemize}
\setlength\itemsep{-0.3em}
\vspace{-0.3em}
  \item \textit{Coverage}: Does the conversation cover the cloze test in the evaluation?
  \item \textit{Appropriateness}: Are the questions properly raised, and appropriate for patient education? 
  \item \textit{Education Outcome}: How do you think the learning experience improves your understanding of discharge instructions?
  \item \textit{Overall}: How do you like the general learning experience considering the above aspects?
\end{itemize}

We report the Mean Reciprocal Rank (MRR) \cite{radev2002evaluating} of each model's final ranking. 
Generally, a higher MRR value implies the evaluators have more preference over an approach. 

\begin{table}[ht]
\setlength{\tabcolsep}{3.9pt}
\renewcommand{\arraystretch}{1.05}
\centering
\begin{scriptsize}
\begin{tabularx}{0.48\textwidth}{X} 

\toprule

You are a physician who wants to evaluate how helpful an AI model is for educating patients. The model asks the patient questions, then verifies the patient's answers, in order to help patients memorize their discharge instructions.

Four evaluation aspects for AI model’s question quality includes:

\emph{Coverage}: Does the conversation cover the cloze test in the evaluation?

\emph{Question Appropriateness}: Are the answers to the questions contained in the discharge instruction?

\emph{Education Outcome}: Do you think the chatbot helps patients understand their discharge instructions?

\emph{Overall}: How do you like the general experience with the chatbot considering the above aspects?

Two evaluation aspects of the AI model’s feedback includes:

\emph{Correctness}: Are the responses from the AI model factually correct?

\emph{Education Potential}: Do the AI model's responses provide helpful information for educating patients?

5-point Likert scale:

1: very low rating

2: low rating

3: neutral or medium rating

4: higher rating

5: very highly rating

The patient's discharge instructions: [\emph{The Patient's Discharge Instruction}]

The conversation between the patient and the AI model: [\emph{The Conversation History}]

Give the 5-point Likert scale of the AI model's question quality (four aspects) and answer feedback (two aspects) one by one. Return the scores as dictionary objects, adhering to the following structure:
{"Coverage": ..., "Question Appropriateness": ...}. 
Please provide your response solely in the dictionary format without including any additional text. \\

\bottomrule

\end{tabularx}
\end{scriptsize}
\vspace{-0.05in}
\caption{Prompt presented to GPT-4 for evaluating the quality of generated questions and answer verification feedback. GPT-4 is expected to output a score on each perspective directly.}
\label{tab:prompt}
\vspace{-0.15in}
\end{table}

\subsection{GPT-4's Automatic Evaluation Settings}
\label{sec:qa_auto_eval}

Following recent practice of applying large language models in evaluating dialogue tasks~\cite{liu2023gpteval}, 
we utilize GPT-4 as the evaluation model to automatically measure the quality of AI generated questions and feedback. 
Similar to patient evaluation in Section~\ref{sec:human_eval_measure}, 
we evaluate the quality of generated questions from the four perspective (i.e. \emph{Coverage}, \emph{Question Appropriateness}, \emph{Education Outcome}  and \emph{Overall}). 
Additionally, we also evaluate the quality of AI models' feedback from two perspectives, i.e. \emph{Correctness} and \emph{Education Potential}. 
Our prompt to the evaluation model is shown in Table~\ref{tab:prompt}. 
We collect evaluation model's responses and report the average score of each perspective.

\subsection{Synthesized Dataset for Evaluation}

Directly presenting real health records to LLMs or participants can lead to data privacy violation.\footnote{\url{https://physionet.org/content/mimiciii/view-dua/1.4/}} Thus, we created 30 synthesized discharge instructions for our human evaluation study. We randomly sampled 30 hospital course notes (a part of EHR data) from the MIMIC-III database, and converted them into synthetic discharge instructions following a neural abstractive summarization method proposed by~\cite{cai-etal-2022-generation}. Our physician collaborators reviewed these synthesized discharge instructions to ensure content validity and anonymity. 

We then apply the various ways (human, GPT, GPT+IE) to created question-answer pairs for these anonymized synthesized data. We demonstrate some sampled discharge instructions and corresponding generated questions in Table~\ref{tab:example}.

\begin{figure*}
    \centering
\includegraphics[width=0.97\textwidth]{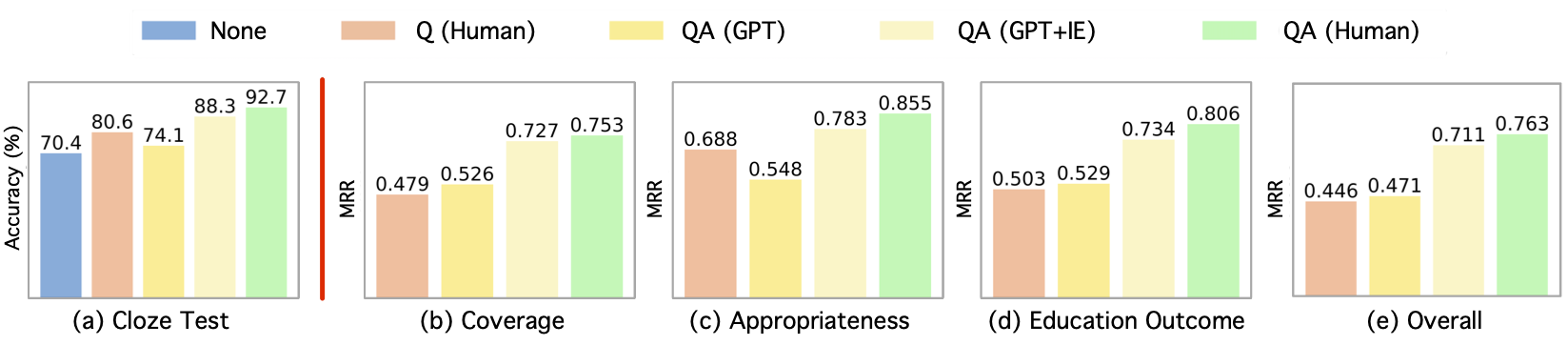}
        \caption{Patient evaluation results, including Cloze Test accuracy and evaluator rankings' MRR scores across four categories (higher is better). The methods are represented with color-coding: \emph{None}-\textcolor{blue_x}{blue}, \emph{Q (Human)}-\textcolor{orange_x}{orange}, \emph{QA (GPT)}-\textcolor{yellow_x}{yellow}, \emph{QA (GPT+IE)}-\textcolor{beige_x}{beige}, \emph{QA (Human)}-\textcolor{green_x}{green}.}
        \label{fig:patient_eval}
        \vspace{-3mm}
\end{figure*}

\subsection{Physician Evaluation Results}

We interview three physician participants with following questions: 
1) Do you think the questions are effective for patients to understand the important info in the discharge instruction? If not, what questions would you ask? 
2) How do you like the questions generated from \textit{GPT} and \textit{GPT+IE}?

Physician participants all believe that \textit{GPT}-generated questions tend to target content that patients do not need to be aware of (e.g., asking why heart attack could cause chest pain is a medical-domain-specific knowledge not suitable for patient's education). 
Sometimes the answers to the GPT-generated questions do not even exist in the discharge instruction.
Take example 1 in Table~\ref{tab:example}, the question asks what the patient should expect in their follow-up visits, but this information is not  mentioned in the discharge instruction.
These qualitative findings may explain why \textit{GPT}-generated questions' are rated by patient participants as low accuracy score in the Cloze Test metric, as well as ranked lower in Coverage, Appropriateness, and Education Outcome in the Section~\ref{patient-result}.

Worth noting, in some cases where the answers are not in the discharge instructions, physician participants actually believe those questions could be useful for patient education. In example 2 in Table~\ref{tab:example}, although the discharge instruction does not contain information on \emph{how to maintain the stent}, physicians still think it is a question they would ask their patients, as it would motivate patients to have better self-managed recovery activities.

For questions generated by \textit{GPT+IE}, 
most questions were perceived by the physicians as appropriate (e.g. example 3).
However, the GPT-IE may still generate improper questions due to errors in the medical event-relation identification. As shown in example 4, the information extraction model identifies the \textit{symptom} ``swelling in your throat'' as a \textit{disease}, which leads to improper questions. 

Physician participants also suggested that some GPT-IE-generated questions lack language fluency. 
As shown in example 5, the generated question seems redundant and can be better rephrased as ``How long do you need to take Prednisone?''

\begin{table}[ht]
    \begin{scriptsize}
    \setlength{\tabcolsep}{2pt}
    \renewcommand{\arraystretch}{1}
    \centering
    \begin{tabularx}{0.48\textwidth}{lX}
    \toprule
    \multirow{2}{*}{1} & \cellcolor{gray!10} \emph{Your symptoms improved and you were discharged to home with close follow-up with your primary care physician and an allergist ...}\\
      & (\textit{\textcolor{red}{GPT}}) \textbf{Q}: When is your follow-up appointment with your primary care physician and allergist and what should you expect during these visits? \\

    \midrule
    \multirow{2}{*}{2} & \cellcolor{gray!10} \emph{We also found that you have a condition called tracheobronchomalacia, which is a blockage of your airways. You had a stent placed in your airway to help keep it open ...} \\
    & (\textit{\textcolor{red}{GPT}}) \textbf{Q}: How should you maintain the stent in your airway? \\
    \midrule
    \multirow{2}{*}{3} & \cellcolor{gray!10} \emph{You were admitted to the hospital with fevers. You were found to have pneumonia, and you were treated with antibiotics ...} \\
    & (\textit{\textcolor{blue}{GPT+IE}}) \textbf{Q1}: What is the cause of your symptom fevers? \textbf{Q2}: What treatment is applied to disease pneumonia? \\
    \midrule
    \multirow{2}{*}{4} & \cellcolor{gray!10} \emph{You were admitted to the hospital with swelling in your throat ...  You were treated with steroids, benadryl, famotidine and epinephrine ...} \\
    & (\textit{\textcolor{blue}{GPT+IE}}) \textbf{Q}: What treatment is applied to disease swelling in your throat? \\
    \midrule
    \multirow{2}{*}{5} & \cellcolor{gray!10} \emph{The following changes have been made to your medications: START Prednisone 40mg daily for 5 days ...} \\
    & (\textit{\textcolor{blue}{GPT+IE}}) \textbf{Q}: What is the recommended duration for taking Prednisone at 40mg daily? \\
    \bottomrule
    \end{tabularx}
    \end{scriptsize}
\caption{Examples of the synthetic discharge instructions and generated questions}
\label{tab:example}
\vspace{-5mm}
\end{table}

\subsection{Patient Evaluation Results}
\label{patient-result}
We summarize the patient evaluation results in Figure \ref{fig:patient_eval}. 
From the (a) Cloze Test chart, we observe that having a chatbot interact with patient participants (regardless of only with \emph{Q} or with both \emph{QA}) can indeed improve their performance over the baseline condition \emph{None}, which suggests our proposed interactive question-answering design is a promising for patient education.
In terms of whether having an answer feedback is helpful or not, the 92.7\% accuracy of \textcolor{green_x}{\emph{QA} (\textit{Human})} significantly outperforms the 80.6\% accuracy performance of \textcolor{orange_x}{\emph{Q}(\textit{Human})}, this implies the importance of validating patients' answers and presenting feedback, thus we decided to always including an answer feedback when conducting further comparison analysis regarding the \textit{GPT} v.s. \textit{GPT+IE} question generation algorithms. 
The result shows that \textcolor{beige_x}{\emph{QA} (\textit{GPT+IE})} 88.3\% achieves higher accuracy than \textcolor{yellow_x}{\emph{QA} (\textit{GPT})} 74.1\%. This demonstrates the improvement by applying enhancements to LLMs for patient education purposes.

The result related to Evaluator Ranking shows (plots (b, c, d, e) in Figure~\ref{fig:patient_eval}): 
1) Considering the Overall ranking of three sets of questions using \emph{QA} interactive approach, \textit{Human} quesions performs better than AI generated questions. This suggests machine-generated questions are still not comparable to human ones. 
2) Comparing the three interactive approaches, we observe \emph{QA} (\textit{Human}) >{}> \emph{Q}(\textit{Human}) > \emph{None}, which is in line with the findings of Cloze Test.
3) In terms of Appropriateness, and Education outcome, \textit{GPT} achieves the lowest ranking. According to our observation, many \textit{GPT}-generated questions ask the evaluators about content not existing in the discharge instruction. As a result, evaluators think the questions are inappropriate and do not help patient education. 
4) \emph{QA}(\textit{GPT+IE}) has higher ranking in Coverage than \emph{QA}(\textit{GPT}). This result is consistent with other recent discussions that incorporate the copying mechanism into LM or LLM by modifying the model structure, loss function, or prompting~\cite{wang2023element, chang2023revisiting, eremeev2023injecting}. \emph{QA}(\textit{Human}) has higher ranking in Coverage than \emph{Q}(\textit{Human}), despite they use the same questions. This suggests much benefit is provided to patients through the answer feedback interaction.

\begin{figure}
    \centering
\includegraphics[width=0.97\linewidth]{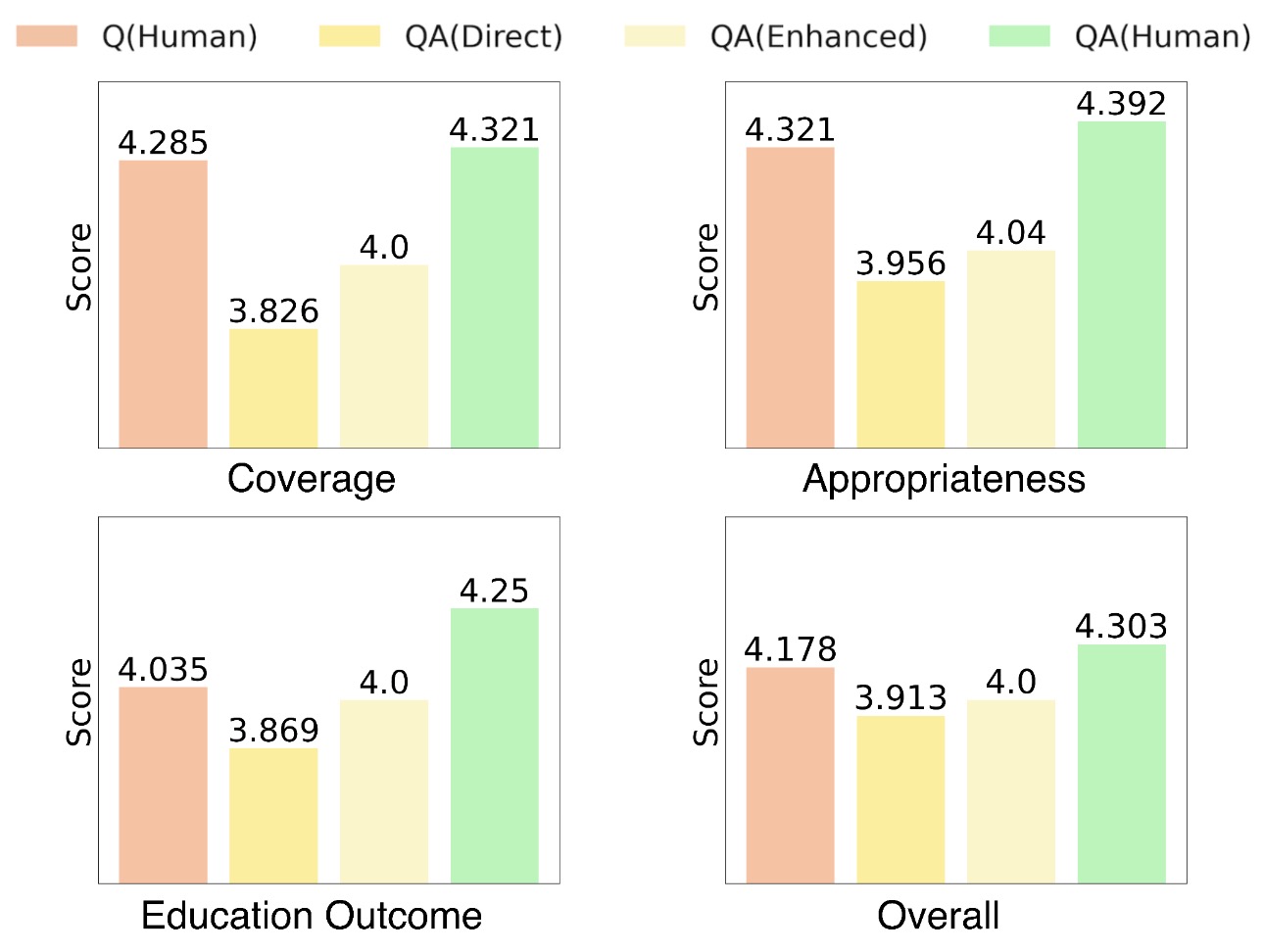}
        \caption{GPT-4's evaluation scores for question quality}
        \label{fig:auto_eval_patient_edu}
        \vspace{-3mm}
\end{figure}

\subsection{GTP-4's Automatic Evaluation Results}

In terms of question quality (as shown in Figure~\ref{fig:auto_eval_patient_edu}), we observe GPT-4's evaluation scores generally follow the same pattern of patient evaluation results, where questions from \emph{Q}(\textit{Enhanced}) are deemed better than \emph{Q}(\textit{Direct}). In addition, we observe the scores of all approaches are close or higher than 4, this implies GPT-4 judges the generated questions are of good quality in four perspectives. In terms of answer verification, as all interactive conditions all share the same verification method, we only present the average \textit{Correctness} and \textit{Education Potential} score. Specifically, GPT-4 gives 4.14 on \textit{Correctness} and 4.01 on \textit{Education Potential}. Both scores are above four, indicating GPT-4 judges feedback from our AI agents' feedback as high quality.

\subsection{Heuristic Evaluation of Conversation Log}
We further conducted a heuristic evaluation to explore the deficiency of AI-generated responses and potential improvements. 
Specifically, we asked an MD student to evaluate the conversation log data of all patient participants.\footnote{The conversation logs are re-used from the patient evaluation described in section~\ref{patient-result}}  
Overall, we collect 192 responses from 30 conversations between the participants mimicking patients and the AI model. 

We ask our MD-background human evaluator to grade each of the AI model's answer feedback, we apply the same evaluation metric, i.e., \textbf{correctness} and \textbf{education potential} as introduced in Section~\ref{sec:qa_auto_eval}. We apply binary coding, i.e. evaluator judge response as positive or negative. 
The positive rate for \textit{Correctness} is 86.4\%, and the positive rate for \textit{Education Potential} is 74.1\%. This suggests that most responses are factually correct and provide helpful information to patients.

Table~\ref{tab:AV_eval} shows some examples of the answer feedback from the chatbot, and we have following design suggestions for future research to improve the quality of the answer feedback: 
1) Most responses are helpful for patients in reviewing their discharge instructions (example 1). 
But, some responses are factually incorrect and may confuse patients. 
The AI model may state that the patient's answer is incorrect or partially correct (example 3), while the patient's response is actually completely correct. 
2) While the responses are generally helpful, they still have a deficiency in providing sufficient and attentive responses in educating patients like a human physician. As shown in example 4, a physician will provide more information about the distinctions between the two medications, including the specific diseases for which they are prescribed.

\begin{table}
    \begin{scriptsize}
    \setlength{\tabcolsep}{2.5pt}
    \renewcommand{\arraystretch}{1.05}
    \centering
    \begin{tabularx}{0.48\textwidth}{lp{7.2cm}}
    \toprule
    \multirow{2}{*}{1} & 
    \textbf{Response helps patients review instructions accurately}
    \newline
    \cellcolor{green!10}
    ``\emph{The following changes were made to your medication regimen. 1. Please take HCTZ 25mg three times a day ...}'' \\
    & \textbf{Bot:} What is the dosage of HCTZ? 
    \newline
    \textbf{User:} 25mg 
    \newline
    \textbf{Bot:} Your answer is partially correct. The dosage of HCTZ is 25mg \underline{three times a day}. \\
    \midrule

    \multirow{2}{*}{2} & 
    \textbf{Response provides information with extra educational value}
    \newline
    \cellcolor{green!10}
    ``\emph{We made the following changes to your medications: -STOPPED Verapamil -STARTED Guaifenesin for airway stent maintenance -INCREASED Nifedipine to 40 mg daily ...}''\\
    & \textbf{Bot:} What medicine is used to treat your disease {low heart rate and high blood pressure}? 
    \newline
    \textbf{User:} guaifensin and nifedipine \newline
    \textbf{Bot:} Your answer is correct. Guaifenesin is used for airway stent maintenance, while Nifedipine is used to lower blood pressure. \\
    \midrule

    \multirow{2}{*}{3} & 
    \textbf{Response is factually incorrect and confuses patient}
    \newline
    \cellcolor{green!10}
    ``\emph{We gave you medications to lower your blood pressure and your breathing improved ...}''\\
    & \textbf{Bot:} What is the result of the medication given to you during your hospital stay? \newline
    \textbf{User:} my blood pressure decreased and my breathing improved \newline
    \textbf{Bot:} The patient's answer is partially correct. The medication helped to improve the patient's breathing. \\
    \midrule

    \multirow{2}{*}{4} & 
    \textbf{Response provides information with little educational value}
    \newline
    \cellcolor{green!10}
    ``\emph{The following changes were made to your medications: STOPPED Losartan CHANGED Lasix to 40mg by mouth twice daily ...}''\\
    & \textbf{Bot:} Should you continue to take Losartan? \newline
    \textbf{User:} no, changed to lasix 
    \newline
    \textbf{Bot:} Your answer is correct. \\
    \bottomrule
    \end{tabularx}
    \end{scriptsize}
\caption{Examples of chatbot's answer feedback.}
\label{tab:AV_eval}
\vspace{-5mm}
\end{table}

\section{Limitations and Ethical Considerations}
\label{sec:ethics}

This study offers valuable insights, but with a few limitations we would like to note.

\vspace{0.04in}
\noindent\textbf{\emph{Biases.}}\quad Large language models trained on vast amounts of text data can pick up biases present in data. For example, they may prefer certain questions related to Aspirin or even associate certain health conditions with specific groups of people. They may also perpetuate misinformation and provide incorrect information. 
In addition, people who participated in our evaluation have different levels of language proficiency and medical background. 
These biases may be mitigated by enhancing model alignment with each individual's background and health literacy level.

\vspace{0.04in}
\noindent\textbf{\emph{Broader Impacts.}}\quad We have performed a preliminary study to educate patients on discharge instructions using interactive question answering. Although we evaluated our system using the MIMIC III dataset, which represents an intensive care unit (ICU) setting, the system should be generalizable to other settings, including perioperative care (from preparation before the surgery to recovery after the surgery), cancer treatment, and chronic condition management.
Our system may help patients receive customized information that is tailored to their individual needs and preferences.

\vspace{0.04in}
\noindent\textbf{\emph{Social Influence.}}\quad Our system has two pillars. First, it is grounded in discharge notes, where we identify important medical events and their relationships that patients should know. Second, it serves an education purpose. For that, we explore the P.E.E.R sequence to prompt the patient, evaluate, extend and ask them to repeat the answer to reinforce their understanding. Additionally, social influence strategies such as small talk, empathy, persuasion can be explored in the future to shape, reinforce, or change a patient’s behavior and promote engagement. 

\vspace{0.04in}
\noindent\textbf{\emph{Privacy Implications.}}\quad LLMs can present privacy concerns in patient education when health records are used, potentially violating the HIPPA regulations. However, in this study, we handle data usage with great care. We conduct all experiments on \emph{open-sourced} real patient data and present an approach to synthetic patient discharge notes. Each synthetic discharge note used in this study has been reviewed by physicians to ensure their validity. We strictly limit our API usage to synthetic data.

\section{Conclusion}
\label{sec:conclusion}

In this study, we present \textbf{\emph{PaniniQA}}, a \textbf{pa}tient-ce\textbf{n}tr\textbf{i}c i\textbf{n}teract\textbf{i}ve question answering system designed to help patients understand and memorize their discharge instructions.  PaniniQA generates educational questions from discharge instructions after identifying salient medical events and event relations. LLMs with prompting is promising for question-answer generation, but sometimes hallucinating. Extensive evaluations highlight the importance of providing answer feedback.

\section*{Acknowledgement}

The authors would like to express sincere gratitude to Center for Biomedical and Health Research in Data Sciences, UMass Lowell which made this research possible.
Hong Yu is supported in part by NIH R01DA056470 and 1R01AG080670, NSF IIS 2124126, and HSR\&D 1I01HX003711-01A1.
Fei Liu is supported in part by National Science Foundation grant IIS-2303678.
The content is solely the responsibility of the authors and does not necessarily represent the official views of the National Institutes of Health, National Science Foundation, and Health Services Research \& Development.

\bibliography{anthology,custom}
\bibliographystyle{acl_natbib}

\end{document}